\documentclass[letterpaper]{article} 
\usepackage[utf8]{inputenc}
\usepackage{newunicodechar}
\newunicodechar{♫}{\textmusicalnote}

\usepackage{aaai2026}  
\usepackage{times}  
\usepackage{helvet}  
\usepackage{courier}  
\usepackage[hyphens]{url}  
\usepackage{graphicx} 
\usepackage{natbib}  
\usepackage{caption} 
\usepackage{algorithm}
\usepackage{algorithmic}

\usepackage{amsthm}
\usepackage{tcolorbox}
\tcbuselibrary{listingsutf8}
\usepackage{listings}

\newtheorem{definition}{Definition}
\newtheorem{theorem}{Theorem}

\usepackage{newfloat}
\usepackage{listings}
\DeclareCaptionStyle{ruled}{labelfont=normalfont,labelsep=colon,strut=off} 
\floatstyle{ruled}
\newfloat{listing}{tb}{lst}{}
\floatname{listing}{Listing}
\title{Beyond Perplexity: Let the Reader Select Retrieval Summaries via Spectrum Projection Score}
\author {
    Zhanghao Hu\textsuperscript{\rm 1},
    Qinglin Zhu\textsuperscript{\rm 1},
    Siya Qi\textsuperscript{\rm 1},
    Yulan He\textsuperscript{\rm 1,2},
    Hanqi Yan\textsuperscript{\rm 1}\thanks{Hanqi Yan and Lin Gui are corresponding authors.},
    Lin Gui\textsuperscript{\rm 1}\footnotemark[1]
}
\affiliations {
    \textsuperscript{\rm 1}King's College London \\
    \textsuperscript{\rm 2}The Alan Turing Institute \\
    \{zhanghao.hu, qinglin.1.zhu, siya.qi,yulan.he, hanqi.1.yan, lin.1.gui\}@kcl.ac.uk \\
}

\usepackage{bibentry}

\usepackage{xspace}
\usepackage{multicol}
\usepackage{amsmath}
\usepackage{comment}
\usepackage{amssymb}
\usepackage{booktabs}
\usepackage{multirow}
\usepackage{makecell}
\usepackage{graphicx}
\usepackage{listings}
\usepackage{enumitem}

\usepackage{pgfplots}
\usepackage{subcaption}

\newcommand{\frameworkname}{xCompress\xspace}
\newcommand{\metricname}{Spectrum Projection Score\xspace}
\newcommand{\shortmetricname}{SPS\xspace}

\begin{document}

\maketitle

\begin{abstract}
Large Language Models (LLMs) have shown improved generation performance through retrieval-augmented generation (RAG) following the retriever-reader paradigm, which supplements model inputs with externally retrieved knowledge. However, prior work often evaluates RAG holistically, assessing the retriever and reader jointly, making it difficult to isolate the true contribution of retrieval, particularly given the prompt sensitivity of LLMs used as readers. We move beyond perplexity and introduce Spectrum Projection Score (SPS), a lightweight and supervision-free metric that allows the reader to gauge the semantic alignment of a retrieved summary with its hidden representation by comparing the area formed by generated tokens from the summary, and the principal directions of subspace in the reader and to measure the relevance. Building on SPS we present \frameworkname, an inference‑time controller framework that dynamically samples, ranks, and compresses retrieval summary candidates. Extensive experiments on five QA benchmarks with four open-sourced LLMs show that SPS not only enhances performance across a range of tasks but also provides a principled perspective on the interaction between retrieval and generation.
\end{abstract}


\begin{links}
    \link{Code}{https://zhanghao-aaai2026-sps.github.io/AAAI2026-SPS/}
\end{links}

\section{Introduction}

Large-context Retrieval-Augmented Generation (RAG) has demonstrated promising capabilities in addressing open-domain question answering tasks \cite{wang-etal-2024-searching,hu-etal-2024-eee}. 
In the standard pipeline, a retriever locates and compresses external evidence with a compressor language model, and a reader generates the final answer from the compressed summary \cite{mialon2023augmented}. A central challenge is to evaluate whether a given summary will actually help the reader answer the question \cite{pmlr-v202-shi23a}, particularly given the reader’s sensitivity to summary variations.  

\begin{figure}[!th]
    \centering
    \includegraphics[width=1\linewidth]{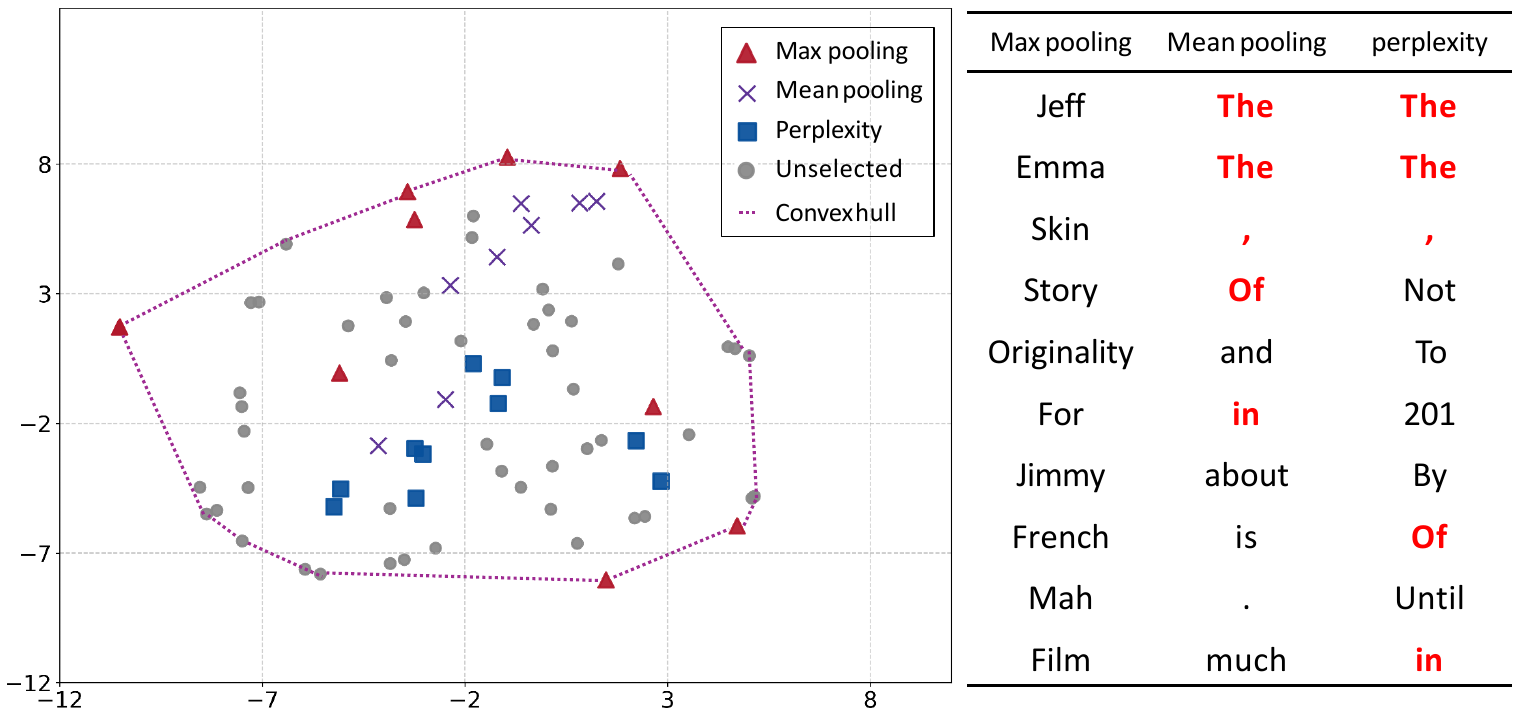}
    \caption{Token selection in the reader’s embedding space. We project summary's token embeddings with t-SNE and compare three selections: nearest to the mean-pooled vector, highest predictive probability, and contributors to max pooling. Mean pooling and perplexity concentrate near the center and favour syntactically frequent tokens. Max pooling emphasises boundary tokens near the convex hull that carry salient semantics.}
    \label{fig:Gau-PCA}
\end{figure}

Therefore, a metric that measures the compatibility of the reader with different input summaries is crucial for better compressed summary generation. Existing measurements, such as token-level perplexity and its long-context variants, or on embedding similarity computed with mean pooling \cite{zhang2025entropyregularized,liu2025nover,chen2024inside}, primarily assess how typical a token sequence is under a language model.
To demonstrate the effects of these measurements, we scatter all tokens from one summary in the reader’s embedding space with t-SNE (Figure~\ref{fig:Gau-PCA}). For this summary, we highlight token embeddings with different selection methods: tokens nearest to the mean-pooled vector, tokens chosen by perplexity-based scoring with the highest predictive probability, and tokens that contribute to max pooling as per-dimension maxima. 
Perplexity and mean pooling favour centrally clustered, low-content tokens (e.g., \textit{the}, \textit{of}, ``,''); by contrast, max pooling surfaces boundary tokens near the convex hull with substantive meaning (e.g., \textit{Jeff}, \textit{Emma}, \textit{French}).
This evidence suggests that representing a sentence by a single centroid or by the most probable tokens fails to capture the shape of the information carried by the sequence.

Given the observation above, we argue that these token-level predictive probability-related embeddings not be able to fully represent the semantic meaning of a text segment, which aims to project text into only one single point in the space and ignore the shape of the distribution. Instead, it should be captured by the collective ``area'' covered by the embeddings of all tokens in the sequence. 
However, two primary challenges arise when adopting this ``area-based'' approach: 1) \textit{Defining the Area:} Formally defining the semantic ``area'' in high-dimensional embedding space is non-trivial. The embedding space is vast, and computing structures like the convex hull that encapsulates all token embeddings can be computationally expensive.
2) \textit{Measuring the Area:} Even if a well-defined area is obtained, measuring its shape and size accurately, which is defined in high-dimensional space, is complex due to potential non-convexity and irregular geometry.

To address these challenges, we propose a novel evaluation metric, the \textbf{Spectrum Projection Score (SPS)}, starting from the concepts of convex hulls and partial order theory, which have been insufficiently explored despite widespread study of max-pooling. SPS leverages max-pooling across token embeddings from the retriever to approximate a semantic ``area'' and applies PCA to identify principal semantic directions (spectrum directions). By aligning this semantic area from the retriever with directions derived from the reader's internal embedding space, even when retriever and reader models differ, SPS quantifies the alignment between the summary embeddings and the reader’s representation, offering a principled measure of semantic confidence.

Building upon SPS, we introduce \textbf{xCompress}, an effective framework that incorporates SPS into test-time sampling strategies. The framework adaptively selects text summaries or embedding summaries optimally aligned with the reader’s parameter space and improves retrieval utility. Furthermore, we enhance efficiency through an adaptive norm-guided filtering strategy, dynamically determining the necessity of sampling for each query, thus maintaining generation quality while reducing computational overhead. We evaluate SPS on five Open Question Answer (Open-QA) datasets using four different large language models. Experimental results show that SPS consistently outperforms existing evaluation baselines across most settings. The main contributions of this paper are as follows:
\begin{itemize}
    \item We propose the Spectrum Projection Score (SPS), a training-free metric that measures summary–reader alignment by projecting a max-pooled envelope of token embeddings onto the reader’s principal subspace and using the residual norm as the score.
    \item We present xCompress, an inference-time controller that samples candidate summaries, ranks them with SPS to select those best aligned with the reader, and applies an adaptive norm-guided filter to control computation.
    \item We empirically validate SPS across five datasets and four state-of-the-art open-sourced LLMs, demonstrating superior performance over established baselines.
\end{itemize}

\section{Related Work}
\paragraph{Summary Compression in RAG.} Recent work on RAG summary focuses on condensing retrieved content into query-relevant representations, primarily through text-to-text summarisation or text-to-embedding conversion \cite{li-etal-2024-refiner,ke-etal-2024-bridging}. Text-to-text approaches generate concise summaries through models trained to distil knowledge from larger language models \cite{yoon-etal-2024-compact}. Alternatively, text-to-embedding methods, such as xRAG \cite{cheng2024xrag}, directly convert retrieved passages into embeddings, concatenating them with query embeddings before processing by the reader. These approaches often yield suboptimal alignment between compressed contexts and the downstream reader's internal representations due to inherent discrepancies in model-specific embedding spaces. In contrast, our framework \frameworkname introduces an inference-time, training-free strategy that aligns retrieval summaries with the reader’s semantic space via SPS.

\paragraph{Perplexity-based Metrics for Text Assessment.} Existing evaluation methods for retrieval-based generation predominantly utilise entropy- or perplexity-based metrics, assessing how well a language model predicts tokens given their preceding context \cite{liu2025nover,yu2025rlpr}. Despite their intuitive appeal, these metrics exhibit fundamental limitations. Primarily, perplexity is sensitive to sequence length \cite{wang2022perplexity}, often emphasising predictable but semantically trivial tokens. Consequently, perplexity-based methods inadequately capture the semantic coherence and relevance critical to retrieval-based question-answering \cite{agarwal2024many,fang2025what}. The latest efforts like SePer \cite{daiseper} seek to assess retrieval utility but rely on human preference signals and do not explicitly model representation-level alignment to task performance.  In contrast, \shortmetricname evaluates summaries by aligning their semantic distribution with the reader’s embedding space using max pooling and convex hull theory, thereby emphasising boundary tokens that carry greater semantic relevance over trivial ones.

\section{Preliminary: Summarise retrieval passages to align with the reader}

\begin{figure}[ht]
    \centering
    \includegraphics[width=1\columnwidth]{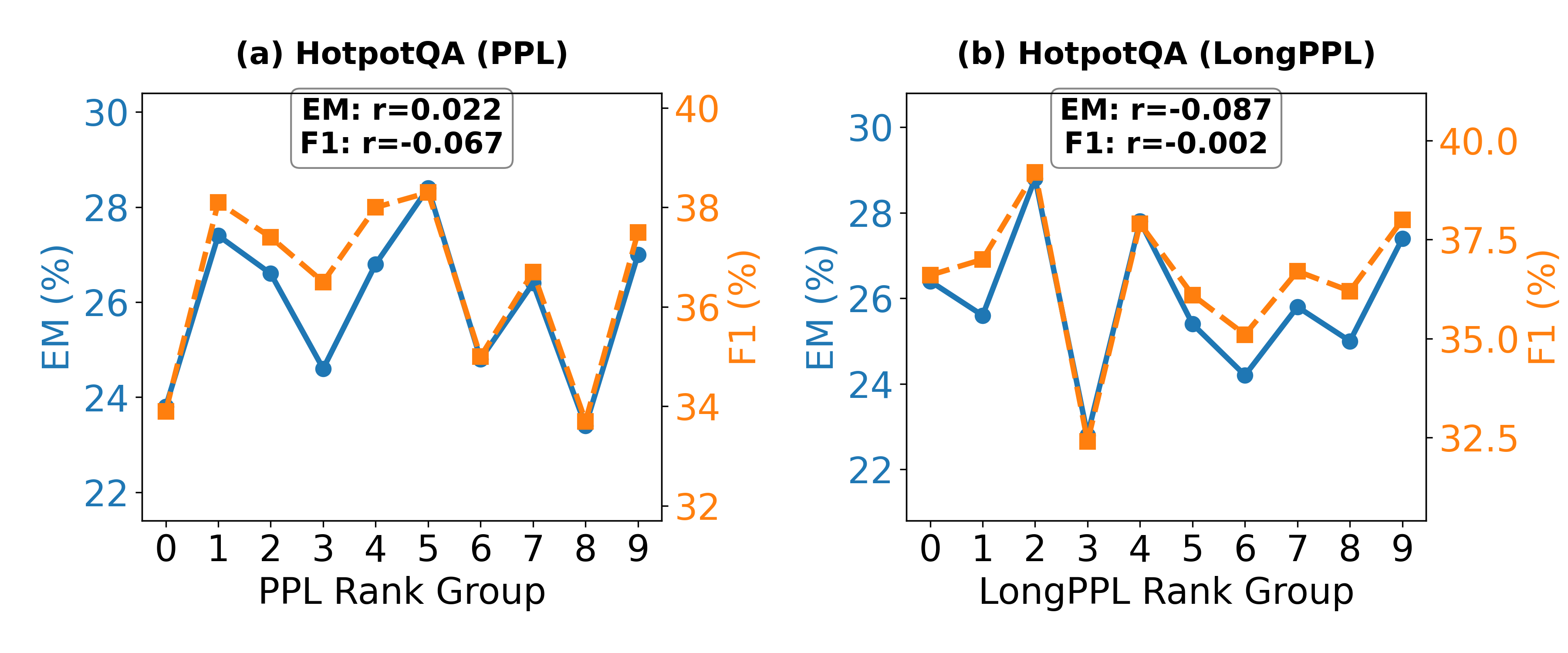}
    \caption{RAG task performances (measured by EM and F1) when feeding summaries with varying PPL (left) and LongPPL (right) to the Reader on the HotpotQA dataset. The low Pearson correlation coefficients ($r$) indicate that both PPL and LongPPL fail to identify a good summary.
    }
    \label{fig:hotpotqa_ppl_longppl_twopanel}
\end{figure}

Retrieval-based generation typically follows a retriever–reader pipeline: documents are retrieved, summarised, and then provided to a generative reader for answer production \cite{lewis2020retrieval,izacardunsupervised,NEURIPS2024_684c59d6}. While summarisation condenses input and highlights salient content \cite{yoon-etal-2024-compact,chengxrag}, prior work has largely overlooked how well a summary \emph{aligns with the reader model's internal representation space}. We therefore focus on the following question: \textit{\textbf{how can we measure the quality of retrieved summaries from the perspective of their compatibility with the reader?}}

A natural proxy is perplexity (PPL), a monotonic transform of sequence likelihood under the reader. Lower PPL indicates that the reader deems a sequence more ``typical'' within its internal language space, suggesting better compatibility. Formally, for a summary token sequence $x=(x_1,\ldots,x_n)$ and a reader parameterised by $P_\theta$, we compute:
\begin{equation}
\mathrm{PPL}_\theta(x)=\exp\!\left(-\frac{1}{n}\sum_{i=1}^{n}\log P_\theta(x_i\mid x_{<i})\right).
\end{equation}

However, our analyses (Figure~\ref{fig:hotpotqa_ppl_longppl_twopanel}a) show a weak association between PPL and downstream QA performance when PPL is used to rank summaries\cite{wang2022perplexity,agarwal2024many}, as well as the alternative implementations such as long PPL \cite{fang2025what}. These observations suggest a structural misalignment: token-level log-likelihood primarily captures \emph{typicality} rather than whether a summary’s \emph{salient semantics} map onto directions that the reader readily encodes and can exploit for answering the query. 

In contrast, we quantify a summary’s semantics by its \emph{coverage} in representation space. Ideally, this is the volume of the convex hull of its token embeddings, but computing high-dimensional hulls is prohibitive. We therefore use a \emph{bounder vector}, which is obtained by elementwise max pooling over token states and forms the minimal axis-aligned enclosure that upper-bounds the token-based convex hull. This summary-level enclosure expands only along informative coordinates as more tokens are added, consistent with the intuition that longer inputs yield greater information. In contrast, low-content tokens (Figure~\ref{fig:Gau-PCA}) typically lie near the centre of the distribution and therefore leave the boundary unchanged. Thus, the bounder provides a simple, stable proxy for a summary’s salient semantics (formal properties in Appendix~B). In summary, in this work, 
the motivation is that: 1) this bounder vector aims to estimate the Essential Supremum, 2) the bounder vector converges to a distribution specific property. 3) Thus, this property of consistent estimation allows the boundary vector to be used as a robust tool to determine if the boundary areas of two different generators, for example, the retriever and the reader in this task, are aligned. In general, this combines sentence-level projection with salient token-level cues, yielding a simple, training-free score that better reflects the utility of retrieved summaries for generation (formalised in Section 4.2). Building on this, the next section defines \metricname, a representation-level score that measures \emph{summary--reader compatibility} by projecting the max-pooling based ``bounder vector'' onto the reader’s principal subspace.

\section{Methodology}

\subsection{Overview and Problem Description}

\begin{figure*}[ht]
    \centering
    \includegraphics[width=1 \textwidth]{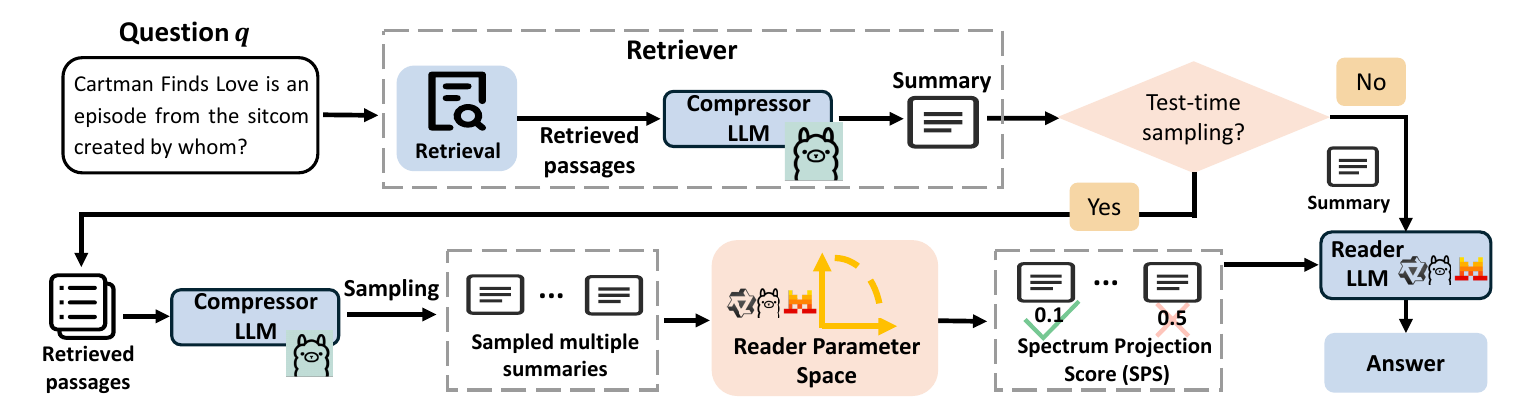}
    \caption{Overview of the \frameworkname framework. Retrieved passages are first compressed into summaries. An adaptive norm-guided filtering mechanism determines whether additional test-time sampling is necessary. If required, multiple summaries are sampled from the compressor LLM and evaluated using the Spectrum Projection Score (SPS). These summaries are first embedded via max-pooling, then projected onto the reader’s principal subspace of its parameter. The summary with the lowest SPS is selected as input to the reader; otherwise, the initial summary is used directly for answer generation.}
    \label{fig:method}
\end{figure*}

\paragraph{Overview.}
We propose \frameworkname, a retrieval-time controller that \emph{evaluates and compresses} retrieved content so it better aligns with the reader LLM (Figure~\ref{fig:method}). In Section~\ref{sec:SPS}, we introduce \metricname, a simple, training-free metric that replaces perplexity-style scoring by assessing summary quality inside the reader’s representation space. In Section~\ref{sec:framework method}, we describe a lightweight test-time sampling that explores both \emph{text-to-text} and \emph{text-to-embedding} compression, ranks candidates with \metricname, and uses an adaptive filtering to decide whether further sampling is needed.

\paragraph{Problem Description.}
We consider the retriever–reader pipeline \cite{lee-etal-2019-latent} for retrieval-based generation. Given a query $q$ and a corpus $\mathcal{D}$, the retriever returns relevant passages $\mathcal{B}$. The top-$N$ passages are compressed into summaries, either in textual (text-to-text) or embedding (text-to-embedding). Each summary candidate is embedded via the reader's penultimate-layer representations, which are subsequently max-pooled. A norm-guided filter then determines if further candidate sampling is necessary. If sampling is triggered, additional summary candidates are generated. Each summary is evaluated by the Spectrum Projection Score (SPS), which measures alignment between the summary’s representation and the reader’s principal embedding subspace. Finally, the candidate with the lowest SPS, or the initial summary if sampling is skipped, is provided to the reader LLM for answer generation.

\subsection{\metricname: Measuring Alignment with the Reader LLM}
\label{sec:SPS}

Retrieval-based generation compresses retrieved documents before feeding them into a reader LLM to reduce context length and foreground salient content \cite{lewis2020retrieval,izacardunsupervised,yoon-etal-2024-compact,chengxrag}. However, entropy- or perplexity-based evaluations are length-biased and only weakly correlated with downstream performance \cite{wang2022perplexity,agarwal2024many,daiseper}, as they capture \emph{typicality} rather than whether a summary’s salient semantics are well represented by the reader’s internal geometry. We therefore seek a representation-level metric that scores a summary by its \emph{compatibility} with the reader.

We introduce the \emph{Spectrum Projection Score} (SPS), which quantifies how well a compressed summary aligns with the reader’s principal representational directions. Let the reader’s representation space be characterised by a matrix $W\in\mathbb{R}^{D\times M}$ (e.g., the input embedding matrix or a bank of hidden states collected from the reader, where $D\times M$ is the corresponding matrix size). We first identify the principal subspace of $W$ through PCA. Specifically, we apply Singular Value Decomposition (SVD): $W=U\Sigma V^\top$, where $U$ and $V$ are left and right singular matrices, $||\cdot||^\top$ is the matrix transpose operation, and $\Sigma$ is the singular value matrix. To retain the principal components by selecting the top $95\%$ eigenvalues in $\Sigma$, noted as $\Sigma_p$, and reconstruct the projection by $P=U\Sigma_p V^\top$, to obtain the reader's core subspace.

Given a retrieval summary, we pass it through the reader and obtain token representations from the penultimate layer; we then apply elementwise max pooling over tokens to form a \emph{salient} summary vector $\mathbf{x}\in\mathbb{R}^{D}$ that preserves boundary features (entities, fact-bearing nouns/adjectives) rather than averaging them away. We define
\begin{equation}
\text{SPS}(\mathbf{x}) = \bigl\| (I - P)\,\mathbf{x} \bigr\|_2,
\end{equation}
\noindent where $I$ is an identity matrix.

Intuitively, this measures how much the main component of bounder vector $\mathbf{x}$ is captured by the reader’s principal subspace: $P\mathbf{x}$ is the in-subspace component, and the residual $(I-P)\mathbf{x}$ quantifies what lies outside. Hence, a smaller $\text{SPS}(\mathbf{x})$ indicates stronger alignment between the summary and the reader’s core representational geometry, making the summary easier for the reader to generate.

\subsection{Test-time Sampling with \metricname}
\label{sec:framework method}
Retrieval-based generation typically follows a sequential retriever–reader pipeline \cite{hu2025beyond}, where an auxiliary LLM compresses retrieved passages into summaries, either textual (text-to-text) \cite{yoon-etal-2024-compact} or embedding-based (text-to-embedding) \cite{cheng2024xrag}, which are then provided to the reader for answer generation. A limitation of this unidirectional flow is that compression is performed without regard to the reader’s internal representational geometry. We instead use the reader’s own representation space to guide compression at test time: we generate a set of summary candidate compressions and score each with \metricname\ (SPS), selecting the summary that best aligns with the reader. 

\paragraph{Test-time Sampling in Text-to-text Compression:} 
In the text-to-text compression paradigm, summaries are typically generated by a compressor LLM before being passed to the reader. To better align these compressed summaries with the reader's embedding space, we propose leveraging the reader's parameters within our SPS metric. Specifically, considering output diversity while maintaining coherence, we adopt stochastic decoding, rather than deterministic methods like greedy or beam search, to produce $K$ diverse summary candidates for each query. Each candidate summary is then evaluated by computing its SPS using the reader’s embedding parameters. Finally, the summary with the lowest SPS, indicating optimal alignment with the reader’s internal representation, is selected as input to the reader LLM for downstream answer generation.

\paragraph{Test-time Sampling in Text-to-embedding Compression.}
Embedding-level compression maps retrieved passages (and the query) directly to a summary embedding via a trained projector \cite{cheng2024xrag}. Because this mapping is deterministic, it offers no native mechanism for sampling diverse candidates. Inspired by soft reasoning with injected noise \cite{hu2025beyond,zhu2025soft}, we introduce \emph{probe-based} stochasticity at test time. Concretely, we sample $N$ small Gaussian probe vectors $\{\mathbf{e}_r\}_{r=1}^N$ and append each to the summary–query embedding before passing the fusion representation through the reader LLM. For each probe, we extract the reader’s penultimate-layer hidden state at the probe position, denoted $\mathbf{h}_r$, and compute a simple diversity score following \citet{hu2025beyond}:
\begin{equation}
    S_{\mathrm{probe}} \;=\; \sum_{i=1}^{p} \bigl(\Delta_{(i)}\bigr)^2,
\end{equation}
where $\Delta_{(i)}$ is the gap between the $i$-th and $(i{+}1)$-th largest elements of $\mathbf{h}_r$. Smaller $S_{\mathrm{probe}}$ indicates stronger semantic deviation from the existing summary–query signal. We retain the $M$ probes with the smallest scores and form $M{+}1$ candidate embedding summaries (the original plus $M$ probed variants). As in the text-to-text case, each candidate is scored with \metricname\ using the reader’s representation space, and the embedding summary with the lowest SPS is selected for answer generation.

\paragraph{Adaptive Sampling via Norm-guided Filtering.}
Sampling multiple candidates improves alignment but is expensive if applied universally. We therefore add a lightweight filter that decides whether further sampling is needed. For the initial summary (text-to-text or text-to-embedding), we compute two proxies in the reader’s latent space: the L2 norm of the \emph{mean-pooled} representation $L2_{mean}$ (captures overall mass) and the L1 norm of the max-pooled representation $L1_{max}$ (captures salient peaks). Their ratio $L2_{mean}/L1_{max}$ serves as a concentration and stability indicator: higher values indicate that the information follows a more skewed distribution, suggesting that the summary is less likely to benefit from additional sampling. In contrast, lower values reflect a more sparse distribution, where further exploration of the alternatives summary may yield additional value (Detailed theoretical discussion is in Appendix B).  
We estimate the threshold by sampling a small subset of the dataset.
In inference, if the ratio exceeds the threshold, we accept the initial summary; otherwise, we perform the sampling-and-selection procedure guided by \metricname. This preserves most of the accuracy gains while substantially reducing computation.

\section{Experiment}

\subsection{Setups.} 
\label{sec:setups}

\paragraph{Evaluation Dataset.} We evaluate our framework on five retrieval-based QA benchmarks: HotpotQA \cite{yang-etal-2018-hotpotqa}, 2WikiMulti-hopQA (2Wiki) \cite{ho-etal-2020-constructing}, Natural Questions (NQ) \cite{kwiatkowski-etal-2019-natural}, TriviaQA (TQA) \cite{joshi-etal-2017-triviaqa}, and Musique \cite{trivedi-etal-2022-musique}. Evaluations are conducted on the development sets, except for TQA, which uses the test set. For NQ, we adopt the original test split with the 21M English Wikipedia dump \cite{karpukhin-etal-2020-dense} as the retrieval corpus. Across all datasets, we follow the data splits and associated document corpora released by \citet{kim2024sure} and \citet{yoon-etal-2024-compact}.

\paragraph{Metrics.} Following prior work \cite{chen2024inside,dai2025seper}, we use the Area Under the Receiver Operating Characteristic curve (AUROC) and Pearson Correlation Coefficient (PCC) to assess the effectiveness of evaluation metrics. AUROC is widely applied to evaluate the measure of uncertainty estimation \cite{chen2024inside}, with higher values indicating better discriminative ability. For retrieval-based generation task performance, we focus on open-domain question answering and report Exact Match (EM) and F1 scores. Following \citet{rajpurkar-etal-2016-squad}, all predictions and gold answers are normalised by lowercasing and removing punctuation to ensure consistency.

\paragraph{Baseline and Models.} We compare \metricname with perplexity, the most common uncertainty-based evaluation metric for large language model predictions \cite{ren2023outofdistribution}, along with its variant, LongPPL, specifically designed to improve performance with long contexts \cite{fang2025what}. For retrieval summary compression in retrieval-based generation, we select one recent method from each compression paradigm: the text-to-text method CompAct \cite{yoon-etal-2024-compact} and the text-to-embedding method xRag \cite{cheng2024xrag}. Additionally, we evaluate various retrieval strategies: 
\begin{itemize}
    \item (1) \textit{Raw Document}, which directly concatenates the top-$k$ retrieved passages; 
    \item (2) \textit{Long-Context LLM Summary}, which uses LLMs to summarise retrieved passages before answer generation, following recent practices \cite{yoon-etal-2024-compact}. 
\end{itemize}
\textbf{Backbone models.} we utilise four open-source LLMs: LLaMA-3.1-8B-Instruct \cite{grattafiori2024llama}, Gemma3-12B-Instruct \cite{team2025gemma}, and Qwen3-8B \cite{yang2025qwen3} for the text-to-text paradigm and select Mistral 7B \cite{jiang2024mixtral} for the text-to-embedding paradigm since the reader model in the text-to-embedding method xRag \cite{cheng2024xrag} is specifically trained alongside its retriever projector, we directly adopt its original reader LLM.

\paragraph{Implementation Details.} For retrieval, we adopt Contriever \cite{izacard2022unsupervised} via the BEIR toolkit \cite{thakur2021beir}. Following \citet{yoon-etal-2024-compact}, we retrieve the top-30 documents for fair comparison. In test-time sampling, we set the temperature to 1.0, apply a repetition penalty of 1.2, and generate five summaries per question to balance diversity and efficiency. For reader LLM generation, we use greedy decoding (temperature = 0.0) to eliminate randomness and ensure reproducibility \cite{sun-etal-2023-chatgpt}(Prompt details are in Appendix A). For norm-guided filtering, we empirically set the threshold as the top-30\% value within the validation set. 

\subsection{Main Experiment}
\begin{table}[htbp]
\centering
\small
\begin{tabular}{llccc}
\toprule
\textbf{Dataset} & \textbf{Metric} & \textbf{PPL} & \textbf{LongPPL} & \textbf{SPS} \\
\midrule
\multirow{3}{*}{HotpotQA}
  & PCC (EM) & 0.022 & -0.087 & \textbf{0.643} \\
  & PCC (F1) & -0.067 & -0.002 & \textbf{0.753} \\
  & AUROC    & 0.504 & 0.495 & \textbf{0.553} \\
\midrule
\multirow{3}{*}{2Wiki}
  & PCC (EM) & -0.318 & -0.065 & \textbf{0.557} \\
  & PCC (F1) & 0.295 & 0.269 & \textbf{0.503} \\
  & AUROC    & 0.487 & 0.482 & \textbf{0.565} \\
\midrule
\multirow{3}{*}{NQ}
  & PCC (EM) & 0.202 & 0.281 & \textbf{0.650} \\
  & PCC (F1) & 0.452 & 0.498 & \textbf{0.628} \\
  & AUROC    & 0.508 & 0.500 & \textbf{0.525} \\
\midrule
\multirow{3}{*}{TQA}
  & PCC (EM) & 0.244 & -0.083 & \textbf{0.563} \\
  & PCC (F1) & 0.127 & -0.210 & \textbf{0.432} \\
  & AUROC    & 0.497 & 0.478 & \textbf{0.531} \\
\midrule
\multirow{3}{*}{Musique}
  & PCC (EM) & 0.182 & -0.186 & \textbf{0.508} \\
  & PCC (F1) & 0.094 & 0.008 & \textbf{0.505} \\
  & AUROC    & 0.443 & 0.488 & \textbf{0.504} \\
\bottomrule
\end{tabular}
\caption{Pearson correlation coefficients (PCC) and AUROC for PPL, LongPPL, and SPS on RAG tasks across five datasets using the LLAMA-3.1-8B-Instruct model as the backbone. SPS consistently achieves the highest correlation with answer quality across all datasets, demonstrating its effectiveness in identifying high-quality summaries.}
\label{tab:metric-correlation-auroc}
\end{table}

\begin{table*}[ht]
\centering
\begin{tabular}{l l c c c c c}
\toprule
\textbf{Model} & \textbf{Method/Dataset} & \textbf{HotpotQA} & \textbf{2WikiMQA} & \textbf{Musique} & \textbf{NQ} & \textbf{TriviaQA} \\
\midrule
\multicolumn{7}{c}{\textbf{Text-to-Text}} \\
\midrule
\multirow{5}{*}{Llama 3.1 8b Ins}
    & Retrieval direct  & 19.6 / 28.85 & 9.4 / 18.22 & 2.0 / 7.57 & 17.4 / 29.58 & 49.4 / 57.64 \\
    & Compact             & 34.0 / 43.17 & 27.2 / 31.83 & 6.6 / 13.76 & 35.2 / 47.49 & 62.4 / 71.25 \\
    & \frameworkname + PPL     & 33.0 / 43.52 & 25.0 / 29.58 & 7.6 / 14.49 & 35.0 / 46.53 & 62.6 / 71.76 \\
    & \frameworkname + LongPPL & 32.0 / 42.48 & 24.2 / 28.16 & 6.8 / 14.71 & 35.6 / 46.78 & 62.4 / 71.72 \\
        & \frameworkname + SPS(ours)     & \textbf{37.6} / \textbf{47.87} & \textbf{29.6} / \textbf{34.21} & \textbf{9.0} / \textbf{17.63} & \textbf{39.4} / \textbf{51.18} & \textbf{65.4} / \textbf{73.11} \\
\midrule
\multirow{4}{*}{Qwen3 8b}
    & Retrieval direct  & 21.4 / 32.29 & 12.2 / 22.19 & 3.8 / 12.49 & 17.8 / 27.70 & 51.8 / 59.64 \\
    & Compact             & 26.8 / 38.84 & 22.2 / 26.69  & 6.0 / 13.36 & 25.8 / 36.50 & 55.4 / 63.22 \\
    & \frameworkname + PPL     & \textbf{29.6} / \textbf{40.86} & 21.8 / 25.63 & 5.0 / 13.48 & \textbf{30.0} / \textbf{45.88} & 54.8 / 63.57 \\
    & \frameworkname + LongPPL & 22.8 / 31.90 & 20.8 / 26.33 & 7.0 / 15.40 & 25.0 / 36.44 & 57.8 / 68.02 \\
        & \frameworkname + SPS(ours)     & 28.8 / 41.84 & \textbf{25.6} / \textbf{30.71} & \textbf{8.6} / \textbf{17.07} & 28.0 / 38.75 & \textbf{59.6} / \textbf{68.84} \\
\midrule
\multirow{5}{*}{Gemma 3 12b Ins}
    & Retrieval direct  & 10.8 / 16.47 & 3.4 / 6.49 & 1.2 / 3.92 & 16.6 / 26.15 & 27.4 / 37.27 \\
    & Compact             & 19.2 / 29.69 & 23.8 / 28.97 & 5.4 / 12.78 & 27.6 / 40.86 & 52.8 / 64.44 \\
    & \frameworkname + PPL     & 19.4 / 31.56 & 23.0 / 28.46 & 5.4 / 13.93 & 28.8 / 39.07 & 52.0 / 64.10 \\
    & \frameworkname + LongPPL & 22.6 / 33.82 & 18.2 / 23.70 & 3.8 / 10.93 & 29.4 / 40.90 & 53.0 / 64.06 \\    
    & \frameworkname + SPS(ours)     & \textbf{25.2} / \textbf{35.66} & \textbf{25.0} / \textbf{29.31} & \textbf{6.4} / \textbf{14.60} & \textbf{31.6} / \textbf{42.27} & \textbf{57.4} / \textbf{65.39} \\
\midrule
\multicolumn{7}{c}{\textbf{Text-to-Embedding}} \\
\midrule
\multirow{3}{*}{Mistral-7b}
    & Retrieval direct  & 1.0 / 8.23 & 1.2 / 11.73 & 0.2 / 3.41 & 1.0 / 5.53 & 2.0 / 14.91 \\
    & xRAG                & 5.2 / 16.63 & 2.2 / 14.09 & 0.4 / 5.69 & 3.0 / 13.56 & 16.0 / 40.09 \\
    & \frameworkname + SPS(ours)     & \textbf{7.6} / \textbf{20.06} & \textbf{2.8} / \textbf{15.82} & \textbf{0.6} / \textbf{6.15} & \textbf{3.8} / \textbf{17.70} & \textbf{29.2} / \textbf{46.76} \\
\bottomrule
\end{tabular}
\caption{EM / F1 (\%) scores across five QA benchmarks using different retrieval and summarisation strategies. \frameworkname with SPS consistently achieves the best performance across models and datasets, demonstrating its effectiveness over perplexity-based metrics and baseline methods in both text-to-text and text-to-embedding paradigms.}
\label{tab:main_results_combined}
\end{table*}

\paragraph{Effectiveness of \metricname.} To evaluate the effectiveness of \metricname in correlating with performance on retrieval-based generation tasks, we conducted experiments comparing it against standard metrics, PPL and LongPPL. Specifically, we generated ten distinct summaries per query using fixed decoding parameters. Each summary was then independently scored using PPL, LongPPL, and our Spectrum Projection Score. To ensure fair comparison, summaries for each query were ranked according to these metric scores, and subsequently grouped into ten ordered bins. We then measured the downstream retrieval-based generation task performance (Exact Match [EM] and F1) for each bin. The correlation between bin rankings and corresponding task performance was quantified using the Pearson Correlation Coefficient (PCC).

Additionally, to further quantify metric discriminative capability, we computed the AUROC scores based on binary correctness (EM=1 as positive, EM=0 as negative). Each pairwise comparison between positive and negative summaries for a given query was used to evaluate whether the metrics correctly identified the better summary or not.

Table \ref{tab:metric-correlation-auroc} demonstrates that both PPL and LongPPL have poor correlation with downstream task performance. Conversely, our Spectrum Projection Score consistently shows significantly stronger correlations, indicating superior effectiveness in distinguishing summary quality relevant to retrieval-based generation.

\paragraph{Effectiveness of \frameworkname.} Table \ref{tab:main_results_combined} summarises the effectiveness of our proposed \frameworkname framework across five QA datasets and two retrieval-based generation paradigms (text-to-text and text-to-embedding). Results demonstrate that incorporating our Spectrum Projection Score (SPS) consistently improves downstream Exact Match (EM) and F1 scores over the baseline compression methods. For instance, on the NQ dataset using the LLAMA 3.1 model, SPS improves performance from 35.2/47.49 (EM/F1) to 39.4/51.18, highlighting SPS's effectiveness in selecting retrieval summaries that align better with the reader model's internal representations. Our framework consistently outperforms perplexity-based (PPL and LongPPL) methods, except in two cases involving the Qwen3 model on the HotpotQA and NQ datasets. In these instances, selecting summaries based on PPL yielded slightly better performance than SPS. We attribute this to the Qwen model's overconfidence on these datasets, possibly due to substantial overlap between its pretraining data and these evaluation datasets, a hypothesis supported by recent literature suggesting dataset contamination or repeated exposure during training \cite{wu2025reasoning}. Further, empirical analysis revealed significantly lower entropy, approximately three times lower, in Qwen's predictions on HotpotQA and NQ compared to LLAMA 3.1, reinforcing that Qwen likely recalls answers directly from memorised content (Experiment details are in the Appendix C.). Consequently, summaries chosen via perplexity reflect familiar, memorised content rather than optimal alignment, paradoxically enhancing performance but reducing generalisability.

\section{Analysis}
\subsection{Why Max Pooling Yields Superior Results?} 
We investigate the impact of different sentence embedding extraction methods, max pooling, mean pooling, and last-token pooling, on our Spectrum Projection Score (SPS) performance. Table \ref{tab:pooling} shows that max pooling consistently achieves superior EM/F1 scores across multiple retrieval-based generation datasets. This suggests max pooling effectively captures salient semantic tokens, whereas mean pooling dilutes semantic signals by averaging, and last-token pooling disproportionately emphasises sentence-end tokens. Hence, max pooling yields embeddings with richer semantic content and better alignment with the LLM’s embedding space, ultimately enhancing downstream task performance.

\begin{table}[htbp]
\centering
\small
\begin{tabular}{lccc}
    \toprule
    \textbf{Dataset} & \textbf{Max Pooling} & \textbf{Mean Pooling} & \textbf{Last Token} \\
    \midrule
    HotpotQA   & \textbf{37.6} / \textbf{47.87}  & 36.2 / 47.65  & 33.6 / 43.37 \\
    2WikiMQA   & \textbf{29.8} / \textbf{34.21}  & 28.2 / 33.63  & 23.6 / 28.10 \\
    Musique    & \textbf{9.2} / \textbf{17.63}   & 7.8 / 15.63   & 5.8 / 10.49 \\
    NQ         & \textbf{39.6} / \textbf{51.18}  & 37.2 / 49.36  & 35.8 / 47.58 \\
    TriviaQA   & \textbf{65.6} / \textbf{73.11}  & 64.2 / 72.62  & 63.0 / 72.24 \\
    \bottomrule
\end{tabular}
\caption{EM / F1 scores (\%) of different pooling strategies for sentence embedding extraction with LLAMA 3.1 across datasets. Max pooling consistently achieves the best.}
\label{tab:pooling}
\end{table}

\subsection{How \metricname Performs with Different Sentence Embeddings?}

\begin{figure}[ht]
    \centering
    \includegraphics[width=\linewidth]
    {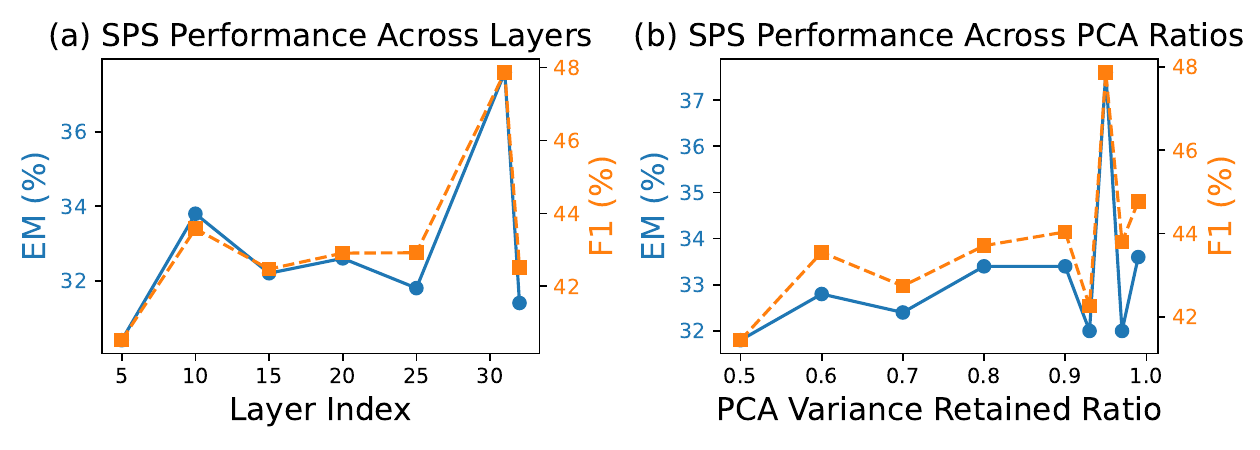}
    \caption{SPS performance under (a) Across LLM layers. (b) Varying PCA retained variance ratios. Optimal results are achieved using embeddings from the penultimate layer and a PCA variance ratio of 0.95.}
    \label{fig:sps_dual_combined}
\end{figure}

\paragraph{Different Layers.} Building on the use of max-pooled sentence embeddings, we examine how SPS behaves across different model layers. As illustrated in Figure \ref{fig:sps_dual_combined} (a), embeddings derived from the penultimate layer consistently yield superior downstream performance compared to embeddings from shallower or the last layers. Specifically, embeddings from earlier layers lack the high-level semantic abstraction necessary for effectively aligning summaries with the model's embedding space, whereas embeddings from the final layer tend to be overly specialised toward token prediction, diminishing their general semantic representativeness. These results empirically confirm that embeddings from the penultimate layer optimally balance semantic abstraction and contextual generalisation, enhancing retrieval quality assessment and thus improving downstream retrieval-based generation performance.

\paragraph{Different PCA Ratio.} We further examine how the variance ratio retained in PCA affects the performance of our \metricname (SPS). As depicted in Figure \ref{fig:sps_dual_combined} (b), performance (EM and F1 scores) peaks when the retained PCA variance ratio is set to 0.95. Lower variance ratios (e.g., 0.50–0.90) fail to preserve sufficient semantic information, resulting in degraded downstream performance. Conversely, excessively high variance ratios (e.g.,0.99) tend to include redundant or noisy dimensions, slightly diluting the semantic representativeness crucial for effective retrieval-summary alignment. Empirically, retaining 95\% variance achieves an optimal balance between semantic richness and dimensional efficiency.

\subsection{Number of Generation Influence.} 
\begin{figure}[htbp]
  \centering
  \includegraphics[width=\linewidth]{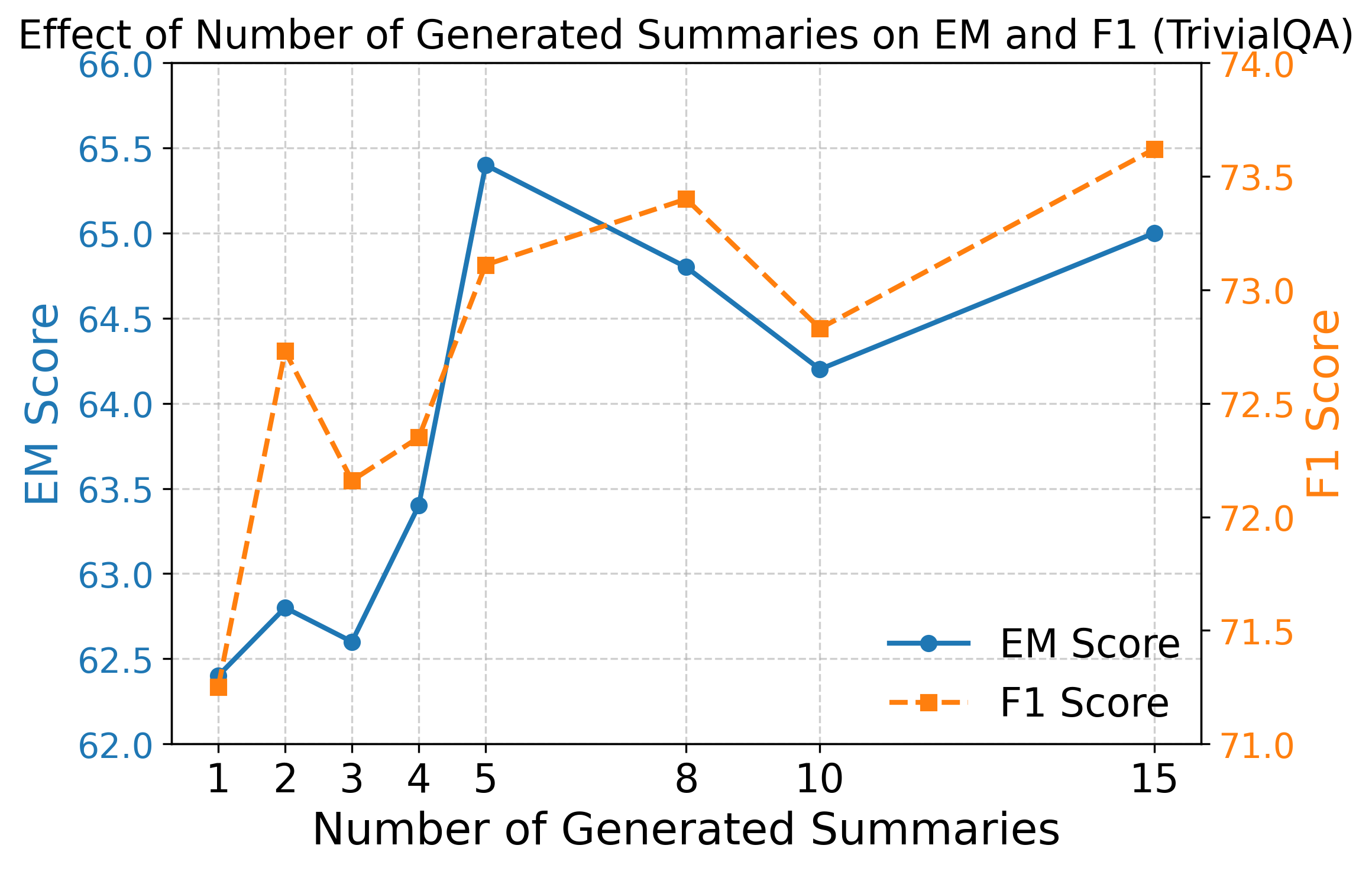}
    \caption{Impact of the number of generated summaries on EM and F1 scores TrivialQA. Performance saturates at five summaries, providing an optimal balance between effectiveness and computational efficiency.}

  \label{fig:generation_number_influence}
\end{figure}
We analyse how the number of generated summaries influences the effectiveness of our proposed Spectrum Projection Score (SPS). As shown in Figure~\ref{fig:generation_number_influence}, \frameworkname surpasses the Compact baseline \cite{yoon-etal-2024-compact} (EM 62.4, F1 71.25) by generating just two summaries, highlighting immediate gains from summary diversity. Performance improves notably up to five summaries (EM 65.4, F1 73.11), beyond which additional summaries yield marginal benefits. Thus, we select five summaries as the optimal balance between performance and computational efficiency.

\section{Conclusion}
We propose the Spectrum Projection Score (SPS), a training-free, representation-level metric that evaluates the semantic alignment between retrieved summaries and the reader model’s internal geometry, moving beyond perplexity and naive similarity as proxies for utility. Building on SPS, we introduce \frameworkname, an inference-time controller that guides summary selection through reader-guided test-time sampling, ranking, and adaptive filtering. Extensive experiments across five QA benchmarks and multiple LLMs demonstrate that SPS not only outperforms perplexity-based baselines in correlation with answer quality and downstream task performance, but also offers a principled, model-agnostic tool for diagnosing and improving retrieval–reader compatibility in RAG systems.

\newpage

\section*{Acknowledgments}
This work was supported by King’s Computational Research, Engineering, and Technology Environment (CREATE) and was supported in part by the UK Engineering and Physical Sciences Research Council (EPSRC) through a Turing AI Fellowship (grant no. EP/V020579/1, EP/V020579/2) and a New Horizons grant (grant no. EP/X019063/1), and KCL’s Impact Acceleration Account (grant no. EP/X525571/1). A PhD studentship from the Chinese Scholarship Council funds Zhanghao Hu. 

\bibliography{aaai2026}

\begin{thebibliography}{43}
\providecommand{\natexlab}[1]{#1}

\bibitem[{Agarwal et~al.(2024)Agarwal, Singh, Zhang, Bohnet, Rosias, Chan, Zhang, Anand, Abbas, Nova et~al.}]{agarwal2024many}
Agarwal, R.; Singh, A.; Zhang, L.; Bohnet, B.; Rosias, L.; Chan, S.; Zhang, B.; Anand, A.; Abbas, Z.; Nova, A.; et~al. 2024.
\newblock Many-shot in-context learning.
\newblock \emph{Advances in Neural Information Processing Systems}, 37: 76930--76966.

\bibitem[{Chen et~al.(2024)Chen, Liu, Chen, Gu, Wu, Tao, Fu, and Ye}]{chen2024inside}
Chen, C.; Liu, K.; Chen, Z.; Gu, Y.; Wu, Y.; Tao, M.; Fu, Z.; and Ye, J. 2024.
\newblock {INSIDE}: {LLM}s' Internal States Retain the Power of Hallucination Detection.
\newblock In \emph{The Twelfth International Conference on Learning Representations}.

\bibitem[{Cheng et~al.(2024{\natexlab{a}})Cheng, Wang, Zhang, Ge, Chen, Wei, Zhang, and Zhao}]{cheng2024xrag}
Cheng, X.; Wang, X.; Zhang, X.; Ge, T.; Chen, S.-Q.; Wei, F.; Zhang, H.; and Zhao, D. 2024{\natexlab{a}}.
\newblock x{RAG}: Extreme Context Compression for Retrieval-augmented Generation with One Token.
\newblock In \emph{The Thirty-eighth Annual Conference on Neural Information Processing Systems}.

\bibitem[{Cheng et~al.(2024{\natexlab{b}})Cheng, Wang, Zhang, Ge, Chen, Wei, Zhang, and Zhao}]{chengxrag}
Cheng, X.; Wang, X.; Zhang, X.; Ge, T.; Chen, S.-Q.; Wei, F.; Zhang, H.; and Zhao, D. 2024{\natexlab{b}}.
\newblock xRAG: Extreme Context Compression for Retrieval-augmented Generation with One Token.
\newblock In \emph{The Thirty-eighth Annual Conference on Neural Information Processing Systems}.

\bibitem[{Dai et~al.(2025{\natexlab{a}})Dai, Xu, Ye, Liu, and Xiong}]{daiseper}
Dai, L.; Xu, Y.; Ye, J.; Liu, H.; and Xiong, H. 2025{\natexlab{a}}.
\newblock SePer: Measure Retrieval Utility Through The Lens Of Semantic Perplexity Reduction.
\newblock In \emph{The Thirteenth International Conference on Learning Representations}.

\bibitem[{Dai et~al.(2025{\natexlab{b}})Dai, Xu, Ye, Liu, and Xiong}]{dai2025seper}
Dai, L.; Xu, Y.; Ye, J.; Liu, H.; and Xiong, H. 2025{\natexlab{b}}.
\newblock SePer: Measure Retrieval Utility Through The Lens Of Semantic Perplexity Reduction.
\newblock In \emph{The Thirteenth International Conference on Learning Representations}.

\bibitem[{Fang et~al.(2025)Fang, Wang, Liu, Zhang, Jegelka, Gao, Ding, and Wang}]{fang2025what}
Fang, L.; Wang, Y.; Liu, Z.; Zhang, C.; Jegelka, S.; Gao, J.; Ding, B.; and Wang, Y. 2025.
\newblock What is Wrong with Perplexity for Long-context Language Modeling?
\newblock In \emph{The Thirteenth International Conference on Learning Representations}.

\bibitem[{Grattafiori et~al.(2024)Grattafiori, Dubey, Jauhri, Pandey, Kadian, Al-Dahle, Letman, Mathur, Schelten, Vaughan et~al.}]{grattafiori2024llama}
Grattafiori, A.; Dubey, A.; Jauhri, A.; Pandey, A.; Kadian, A.; Al-Dahle, A.; Letman, A.; Mathur, A.; Schelten, A.; Vaughan, A.; et~al. 2024.
\newblock The llama 3 herd of models.
\newblock \emph{arXiv preprint arXiv:2407.21783}.

\bibitem[{Ho et~al.(2020)Ho, Duong~Nguyen, Sugawara, and Aizawa}]{ho-etal-2020-constructing}
Ho, X.; Duong~Nguyen, A.-K.; Sugawara, S.; and Aizawa, A. 2020.
\newblock Constructing A Multi-hop {QA} Dataset for Comprehensive Evaluation of Reasoning Steps.
\newblock In Scott, D.; Bel, N.; and Zong, C., eds., \emph{Proceedings of the 28th International Conference on Computational Linguistics}, 6609--6625. Barcelona, Spain (Online): International Committee on Computational Linguistics.

\bibitem[{Hu et~al.(2025)Hu, Yan, Zhu, Shen, He, and Gui}]{hu2025beyond}
Hu, Z.; Yan, H.; Zhu, Q.; Shen, Z.; He, Y.; and Gui, L. 2025.
\newblock Beyond Prompting: An Efficient Embedding Framework for Open-Domain Question Answering.
\newblock \emph{arXiv preprint arXiv:2503.01606}.

\bibitem[{Hu et~al.(2024)Hu, Yang, Xu, Qiu, and Chen}]{hu-etal-2024-eee}
Hu, Z.; Yang, Y.; Xu, J.; Qiu, Y.; and Chen, P. 2024.
\newblock {EEE}-{QA}: Exploring Effective and Efficient Question-Answer Representations.
\newblock In Calzolari, N.; Kan, M.-Y.; Hoste, V.; Lenci, A.; Sakti, S.; and Xue, N., eds., \emph{Proceedings of the 2024 Joint International Conference on Computational Linguistics, Language Resources and Evaluation (LREC-COLING 2024)}, 5520--5525. Torino, Italia: ELRA and ICCL.

\bibitem[{Izacard et~al.(2022{\natexlab{a}})Izacard, Caron, Hosseini, Riedel, Bojanowski, Joulin, and Grave}]{izacardunsupervised}
Izacard, G.; Caron, M.; Hosseini, L.; Riedel, S.; Bojanowski, P.; Joulin, A.; and Grave, E. 2022{\natexlab{a}}.
\newblock Unsupervised Dense Information Retrieval with Contrastive Learning.
\newblock \emph{Transactions on Machine Learning Research}.

\bibitem[{Izacard et~al.(2022{\natexlab{b}})Izacard, Caron, Hosseini, Riedel, Bojanowski, Joulin, and Grave}]{izacard2022unsupervised}
Izacard, G.; Caron, M.; Hosseini, L.; Riedel, S.; Bojanowski, P.; Joulin, A.; and Grave, E. 2022{\natexlab{b}}.
\newblock Unsupervised Dense Information Retrieval with Contrastive Learning.
\newblock \emph{Transactions on Machine Learning Research}.

\bibitem[{Jiang et~al.(2024)Jiang, Sablayrolles, Roux, Mensch, Savary, Bamford, Chaplot, Casas, Hanna, Bressand et~al.}]{jiang2024mixtral}
Jiang, A.~Q.; Sablayrolles, A.; Roux, A.; Mensch, A.; Savary, B.; Bamford, C.; Chaplot, D.~S.; Casas, D. d.~l.; Hanna, E.~B.; Bressand, F.; et~al. 2024.
\newblock Mixtral of experts.
\newblock \emph{arXiv preprint arXiv:2401.04088}.

\bibitem[{Joshi et~al.(2017)Joshi, Choi, Weld, and Zettlemoyer}]{joshi-etal-2017-triviaqa}
Joshi, M.; Choi, E.; Weld, D.; and Zettlemoyer, L. 2017.
\newblock {T}rivia{QA}: A Large Scale Distantly Supervised Challenge Dataset for Reading Comprehension.
\newblock In Barzilay, R.; and Kan, M.-Y., eds., \emph{Proceedings of the 55th Annual Meeting of the Association for Computational Linguistics (Volume 1: Long Papers)}, 1601--1611. Vancouver, Canada: Association for Computational Linguistics.

\bibitem[{Karpukhin et~al.(2020)Karpukhin, Oguz, Min, Lewis, Wu, Edunov, Chen, and Yih}]{karpukhin-etal-2020-dense}
Karpukhin, V.; Oguz, B.; Min, S.; Lewis, P.; Wu, L.; Edunov, S.; Chen, D.; and Yih, W.-t. 2020.
\newblock Dense Passage Retrieval for Open-Domain Question Answering.
\newblock In Webber, B.; Cohn, T.; He, Y.; and Liu, Y., eds., \emph{Proceedings of the 2020 Conference on Empirical Methods in Natural Language Processing (EMNLP)}, 6769--6781. Online: Association for Computational Linguistics.

\bibitem[{Karvonen et~al.(2025)Karvonen, Rager, Lin, Tigges, Bloom, Chanin, Lau, Farrell, McDougall, Ayonrinde et~al.}]{karvonen2025saebench}
Karvonen, A.; Rager, C.; Lin, J.; Tigges, C.; Bloom, J.; Chanin, D.; Lau, Y.-T.; Farrell, E.; McDougall, C.; Ayonrinde, K.; et~al. 2025.
\newblock Saebench: A comprehensive benchmark for sparse autoencoders in language model interpretability.
\newblock \emph{arXiv preprint arXiv:2503.09532}.

\bibitem[{Ke et~al.(2024)Ke, Kong, Li, Zhang, Mei, and Bendersky}]{ke-etal-2024-bridging}
Ke, Z.; Kong, W.; Li, C.; Zhang, M.; Mei, Q.; and Bendersky, M. 2024.
\newblock Bridging the Preference Gap between Retrievers and {LLM}s.
\newblock In Ku, L.-W.; Martins, A.; and Srikumar, V., eds., \emph{Proceedings of the 62nd Annual Meeting of the Association for Computational Linguistics (Volume 1: Long Papers)}, 10438--10451. Bangkok, Thailand: Association for Computational Linguistics.

\bibitem[{Kim et~al.(2024)Kim, Nam, Mo, Park, Lee, Seo, Ha, and Shin}]{kim2024sure}
Kim, J.; Nam, J.; Mo, S.; Park, J.; Lee, S.-W.; Seo, M.; Ha, J.-W.; and Shin, J. 2024.
\newblock SuRe: Summarizing Retrievals using Answer Candidates for Open-domain {QA} of {LLM}s.
\newblock In \emph{The Twelfth International Conference on Learning Representations}.

\bibitem[{Kwiatkowski et~al.(2019)Kwiatkowski, Palomaki, Redfield, Collins, Parikh, Alberti, Epstein, Polosukhin, Devlin, Lee, Toutanova, Jones, Kelcey, Chang, Dai, Uszkoreit, Le, and Petrov}]{kwiatkowski-etal-2019-natural}
Kwiatkowski, T.; Palomaki, J.; Redfield, O.; Collins, M.; Parikh, A.; Alberti, C.; Epstein, D.; Polosukhin, I.; Devlin, J.; Lee, K.; Toutanova, K.; Jones, L.; Kelcey, M.; Chang, M.-W.; Dai, A.~M.; Uszkoreit, J.; Le, Q.; and Petrov, S. 2019.
\newblock Natural Questions: A Benchmark for Question Answering Research.
\newblock \emph{Transactions of the Association for Computational Linguistics}, 7: 452--466.

\bibitem[{Lee, Chang, and Toutanova(2019)}]{lee-etal-2019-latent}
Lee, K.; Chang, M.-W.; and Toutanova, K. 2019.
\newblock Latent Retrieval for Weakly Supervised Open Domain Question Answering.
\newblock In Korhonen, A.; Traum, D.; and M{\`a}rquez, L., eds., \emph{Proceedings of the 57th Annual Meeting of the Association for Computational Linguistics}, 6086--6096. Florence, Italy: Association for Computational Linguistics.

\bibitem[{Lewis et~al.(2020)Lewis, Perez, Piktus, Petroni, Karpukhin, Goyal, K{\"u}ttler, Lewis, Yih, Rockt{\"a}schel et~al.}]{lewis2020retrieval}
Lewis, P.; Perez, E.; Piktus, A.; Petroni, F.; Karpukhin, V.; Goyal, N.; K{\"u}ttler, H.; Lewis, M.; Yih, W.-t.; Rockt{\"a}schel, T.; et~al. 2020.
\newblock Retrieval-augmented generation for knowledge-intensive nlp tasks.
\newblock \emph{Advances in neural information processing systems}, 33: 9459--9474.

\bibitem[{Li et~al.(2024)Li, Hu, Liu, Zheng, Huang, and Xiong}]{li-etal-2024-refiner}
Li, Z.; Hu, X.; Liu, A.; Zheng, K.; Huang, S.; and Xiong, H. 2024.
\newblock $\textit{Refiner}$: Restructure Retrieved Content Efficiently to Advance Question-Answering Capabilities.
\newblock In Al-Onaizan, Y.; Bansal, M.; and Chen, Y.-N., eds., \emph{Findings of the Association for Computational Linguistics: EMNLP 2024}, 8548--8572. Miami, Florida, USA: Association for Computational Linguistics.

\bibitem[{Liu et~al.(2025)Liu, Qi, Wang, Qian, Du, and He}]{liu2025nover}
Liu, W.; Qi, S.; Wang, X.; Qian, C.; Du, Y.; and He, Y. 2025.
\newblock NOVER: Incentive Training for Language Models via Verifier-Free Reinforcement Learning.
\newblock \emph{arXiv preprint arXiv:2505.16022}.

\bibitem[{Mialon et~al.(2023)Mialon, Dessi, Lomeli, Nalmpantis, Pasunuru, Raileanu, Roziere, Schick, Dwivedi-Yu, Celikyilmaz, Grave, LeCun, and Scialom}]{mialon2023augmented}
Mialon, G.; Dessi, R.; Lomeli, M.; Nalmpantis, C.; Pasunuru, R.; Raileanu, R.; Roziere, B.; Schick, T.; Dwivedi-Yu, J.; Celikyilmaz, A.; Grave, E.; LeCun, Y.; and Scialom, T. 2023.
\newblock Augmented Language Models: a Survey.
\newblock \emph{Transactions on Machine Learning Research}.
\newblock Survey Certification.

\bibitem[{Ng et~al.(2011)}]{ng2011sparse}
Ng, A.; et~al. 2011.
\newblock Sparse autoencoder.
\newblock \emph{CS294A Lecture notes}, 72(2011): 1--19.

\bibitem[{Qiu et~al.(2024)Qiu, Zhao, Ziser, Korhonen, Ponti, and Cohen}]{NEURIPS2024_684c59d6}
Qiu, Y.; Zhao, Z.; Ziser, Y.; Korhonen, A.; Ponti, E.~M.; and Cohen, S.~B. 2024.
\newblock Spectral Editing of Activations for Large Language Model Alignment.
\newblock In Globerson, A.; Mackey, L.; Belgrave, D.; Fan, A.; Paquet, U.; Tomczak, J.; and Zhang, C., eds., \emph{Advances in Neural Information Processing Systems}, volume~37, 56958--56987. Curran Associates, Inc.

\bibitem[{Rajpurkar et~al.(2016)Rajpurkar, Zhang, Lopyrev, and Liang}]{rajpurkar-etal-2016-squad}
Rajpurkar, P.; Zhang, J.; Lopyrev, K.; and Liang, P. 2016.
\newblock {SQ}u{AD}: 100,000+ Questions for Machine Comprehension of Text.
\newblock In Su, J.; Duh, K.; and Carreras, X., eds., \emph{Proceedings of the 2016 Conference on Empirical Methods in Natural Language Processing}, 2383--2392. Austin, Texas: Association for Computational Linguistics.

\bibitem[{Ren et~al.(2023)Ren, Luo, Zhao, Krishna, Saleh, Lakshminarayanan, and Liu}]{ren2023outofdistribution}
Ren, J.; Luo, J.; Zhao, Y.; Krishna, K.; Saleh, M.; Lakshminarayanan, B.; and Liu, P.~J. 2023.
\newblock Out-of-Distribution Detection and Selective Generation for Conditional Language Models.
\newblock In \emph{The Eleventh International Conference on Learning Representations}.

\bibitem[{Shi et~al.(2023)Shi, Chen, Misra, Scales, Dohan, Chi, Sch\"{a}rli, and Zhou}]{pmlr-v202-shi23a}
Shi, F.; Chen, X.; Misra, K.; Scales, N.; Dohan, D.; Chi, E.~H.; Sch\"{a}rli, N.; and Zhou, D. 2023.
\newblock Large Language Models Can Be Easily Distracted by Irrelevant Context.
\newblock In Krause, A.; Brunskill, E.; Cho, K.; Engelhardt, B.; Sabato, S.; and Scarlett, J., eds., \emph{Proceedings of the 40th International Conference on Machine Learning}, volume 202 of \emph{Proceedings of Machine Learning Research}, 31210--31227. PMLR.

\bibitem[{Sun et~al.(2023)Sun, Yan, Ma, Wang, Ren, Chen, Yin, and Ren}]{sun-etal-2023-chatgpt}
Sun, W.; Yan, L.; Ma, X.; Wang, S.; Ren, P.; Chen, Z.; Yin, D.; and Ren, Z. 2023.
\newblock Is {C}hat{GPT} Good at Search? Investigating Large Language Models as Re-Ranking Agents.
\newblock In Bouamor, H.; Pino, J.; and Bali, K., eds., \emph{Proceedings of the 2023 Conference on Empirical Methods in Natural Language Processing}, 14918--14937. Singapore: Association for Computational Linguistics.

\bibitem[{Team et~al.(2025)Team, Kamath, Ferret, Pathak, Vieillard, Merhej, Perrin, Matejovicova, Ram{\'e}, Rivi{\`e}re et~al.}]{team2025gemma}
Team, G.; Kamath, A.; Ferret, J.; Pathak, S.; Vieillard, N.; Merhej, R.; Perrin, S.; Matejovicova, T.; Ram{\'e}, A.; Rivi{\`e}re, M.; et~al. 2025.
\newblock Gemma 3 technical report.
\newblock \emph{arXiv preprint arXiv:2503.19786}.

\bibitem[{Thakur et~al.(2021)Thakur, Reimers, R{\"u}ckl{\'e}, Srivastava, and Gurevych}]{thakur2021beir}
Thakur, N.; Reimers, N.; R{\"u}ckl{\'e}, A.; Srivastava, A.; and Gurevych, I. 2021.
\newblock {BEIR}: A Heterogeneous Benchmark for Zero-shot Evaluation of Information Retrieval Models.
\newblock In \emph{Thirty-fifth Conference on Neural Information Processing Systems Datasets and Benchmarks Track (Round 2)}.

\bibitem[{Trivedi et~al.(2022)Trivedi, Balasubramanian, Khot, and Sabharwal}]{trivedi-etal-2022-musique}
Trivedi, H.; Balasubramanian, N.; Khot, T.; and Sabharwal, A. 2022.
\newblock ♫ {M}u{S}i{Q}ue: Multihop Questions via Single-hop Question Composition.
\newblock \emph{Transactions of the Association for Computational Linguistics}, 10: 539--554.

\bibitem[{Wang et~al.(2024)Wang, Wang, Gao, Zhang, Wu, Xu, Shi, Wang, Li, Qian, Yin, Lv, Zheng, and Huang}]{wang-etal-2024-searching}
Wang, X.; Wang, Z.; Gao, X.; Zhang, F.; Wu, Y.; Xu, Z.; Shi, T.; Wang, Z.; Li, S.; Qian, Q.; Yin, R.; Lv, C.; Zheng, X.; and Huang, X. 2024.
\newblock Searching for Best Practices in Retrieval-Augmented Generation.
\newblock In Al-Onaizan, Y.; Bansal, M.; and Chen, Y.-N., eds., \emph{Proceedings of the 2024 Conference on Empirical Methods in Natural Language Processing}, 17716--17736. Miami, Florida, USA: Association for Computational Linguistics.

\bibitem[{Wang et~al.(2022)Wang, Deng, Sun, and Meng}]{wang2022perplexity}
Wang, Y.; Deng, J.; Sun, A.; and Meng, X. 2022.
\newblock Perplexity from plm is unreliable for evaluating text quality.
\newblock \emph{arXiv preprint arXiv:2210.05892}.

\bibitem[{Wu et~al.(2025)Wu, Zhang, Dong, Xi, Zhao, Jin, Fan, Zhou, Fu, Liu et~al.}]{wu2025reasoning}
Wu, M.; Zhang, Z.; Dong, Q.; Xi, Z.; Zhao, J.; Jin, S.; Fan, X.; Zhou, Y.; Fu, Y.; Liu, Q.; et~al. 2025.
\newblock Reasoning or Memorization? Unreliable Results of Reinforcement Learning Due to Data Contamination.
\newblock \emph{arXiv preprint arXiv:2507.10532}.

\bibitem[{Yang et~al.(2025)Yang, Li, Yang, Zhang, Hui, Zheng, Yu, Gao, Huang, Lv et~al.}]{yang2025qwen3}
Yang, A.; Li, A.; Yang, B.; Zhang, B.; Hui, B.; Zheng, B.; Yu, B.; Gao, C.; Huang, C.; Lv, C.; et~al. 2025.
\newblock Qwen3 technical report.
\newblock \emph{arXiv preprint arXiv:2505.09388}.

\bibitem[{Yang et~al.(2018)Yang, Qi, Zhang, Bengio, Cohen, Salakhutdinov, and Manning}]{yang-etal-2018-hotpotqa}
Yang, Z.; Qi, P.; Zhang, S.; Bengio, Y.; Cohen, W.; Salakhutdinov, R.; and Manning, C.~D. 2018.
\newblock {H}otpot{QA}: A Dataset for Diverse, Explainable Multi-hop Question Answering.
\newblock In Riloff, E.; Chiang, D.; Hockenmaier, J.; and Tsujii, J., eds., \emph{Proceedings of the 2018 Conference on Empirical Methods in Natural Language Processing}, 2369--2380. Brussels, Belgium: Association for Computational Linguistics.

\bibitem[{Yoon et~al.(2024)Yoon, Lee, Hwang, Jeong, and Kang}]{yoon-etal-2024-compact}
Yoon, C.; Lee, T.; Hwang, H.; Jeong, M.; and Kang, J. 2024.
\newblock {C}omp{A}ct: Compressing Retrieved Documents Actively for Question Answering.
\newblock In Al-Onaizan, Y.; Bansal, M.; and Chen, Y.-N., eds., \emph{Proceedings of the 2024 Conference on Empirical Methods in Natural Language Processing}, 21424--21439. Miami, Florida, USA: Association for Computational Linguistics.

\bibitem[{Yu et~al.(2025)Yu, Ji, Wang, Yao, Wang, Cui, Yuan, Ding, Yao, Liu et~al.}]{yu2025rlpr}
Yu, T.; Ji, B.; Wang, S.; Yao, S.; Wang, Z.; Cui, G.; Yuan, L.; Ding, N.; Yao, Y.; Liu, Z.; et~al. 2025.
\newblock RLPR: Extrapolating RLVR to General Domains without Verifiers.
\newblock \emph{arXiv preprint arXiv:2506.18254}.

\bibitem[{Zhang et~al.(2025)Zhang, Wang, Diao, Lin, Pan, Dong, Zhang, Molchanov, and Zhang}]{zhang2025entropyregularized}
Zhang, H.; Wang, P.; Diao, S.; Lin, Y.; Pan, R.; Dong, H.; Zhang, D.; Molchanov, P.; and Zhang, T. 2025.
\newblock Entropy-Regularized Process Reward Model.
\newblock \emph{Transactions on Machine Learning Research}.

\bibitem[{Zhu et~al.(2025)Zhu, Zhao, Yan, He, Chen, and Gui}]{zhu2025soft}
Zhu, Q.; Zhao, R.; Yan, H.; He, Y.; Chen, Y.; and Gui, L. 2025.
\newblock Soft Reasoning: Navigating Solution Spaces in Large Language Models through Controlled Embedding Exploration.
\newblock In \emph{Forty-second International Conference on Machine Learning}.

\end{thebibliography}

\clearpage

\appendix

\section{A. Prompt Design}
In this section, we present our prompts used for the experiments in Section \ref{sec:setups}.
In Listing 1, we present the prompt $p_{compressor}$, which is used to generate summaries from the given question and $N$ retrieved passages.

\begin{table}[!htbp]
\centering
\small
\label{tab:prompt_multi_generate}
\begin{tcolorbox}[
    colback=gray!5,
    colframe=black,
    width=\linewidth,  
    boxrule=0.5pt,
    arc=2pt,
    outer arc=2pt,
    title={\textbf{Prompt for Compressor LLM}},
    fonttitle=\bfseries
]
\textbf{Instruct}\\
        1. Generate a summary of source documents to answer the question. Ensure the summary is under 200 words and does not include any pronouns. DO NOT make assumptions or attempt to answer the question; your job is to summarise only.\\
        2. Evaluate the summary based solely on the information of it, without any additional background context.\\
        \\
        \textbf{Question}: \{question\} \\
        \\
        \textbf{Source documents}: \{document\_input\} \\
        \\
        \textbf{Summary}:

\end{tcolorbox}
\end{table}

In Listing 2, we present the prompt $p_{reader}$, which is used to generate answers from the given question and summary.

\begin{table}[!htbp]
\centering
\small
\label{tab:prompt_multi_generate}
\begin{tcolorbox}[
    colback=gray!5,
    colframe=black,
    width=\linewidth,  
    boxrule=0.5pt,
    arc=2pt,
    outer arc=2pt,
    title={\textbf{Prompt for Reader LLM}},
    fonttitle=\bfseries
]
Write a high-quality answer for the given question using only the provided search results (some of which might be irrelevant).
\end{tcolorbox}
\end{table}

In Listing 3, we present the prompt $p_{table1}$, which is used to generate the summary from the given question for the table 1 experiment.

\begin{table}[!htbp]
\centering
\small
\label{tab:prompt_multi_generate}
\begin{tcolorbox}[
    colback=gray!5,
    colframe=black,
    width=\linewidth,  
    boxrule=0.5pt,
    arc=2pt,
    outer arc=2pt,
    title={\textbf{Prompt for generating the group of summary in Experiment Table 1}},
    fonttitle=\bfseries
]
\textbf{Instruct}

Your job is to act as a professional writer. You will write a good-quality passage that can support the given prediction about the question only based on the information in the provided supporting passages.

Now, let's start. After you write, please write [DONE] to indicate you are done. Do not write a prefix (e.g., "Response:") while writing a passage.\\

\textbf{Question}: {example['question']}\\

\textbf{Source documents}: {document\_input}\\

\textbf{Summary}:"
\end{tcolorbox}
\end{table}

\section{B. Theoretical discussion}
\label{appendix:norm theoretical discussion}

In this section, we would like to discuss the theoretical intuition to support the methodology. 
\subsection{Bounder and the order}

We first define the concept of a bounded vector and introduce the corresponding notion of hyper-rectangle order. We then prove that the hyper-rectangle order constitutes a partial order. Furthermore, we show that if one convex hull in a high-dimensional vector space encloses another, their corresponding bounded vectors also satisfy the hyper-rectangle order.

\begin{definition}
\textbf{Bounder vector}: Given a sequence $\mathbf{x} = (\mathbf{x}_1, \ldots, \mathbf{x}_m)$, where each $\mathbf{x}_i$ is an $n$-dimensional vector, a vector $\mathbf{M} \in \mathbb{R}^n$ is called a \emph{bounder vector} of $\mathbf{x}$ if, for every $\mathbf{x}_i$ and for every dimension $k \in \{1, \ldots, n\}$, it holds that $M^k \geq x_i^k$, where $M^k$ and $x_i^k$ denote the $k$-th components of $\mathbf{M}$ and $\mathbf{x}_i$, respectively.
\end{definition}

\noindent Obviously, for any token sequence, the max pooling result is a bounder vector of the sequence. Then, we can define a partial order based on the bounder vector.

\begin{definition}
\textbf{Hyper-rectangle order}: Given two sequences $\mathbf{x} = (\mathbf{x}_1, \ldots, \mathbf{x}_m)$ and $\mathbf{y} = (\mathbf{y}_1, \ldots, \mathbf{y}_l)$, where each $\mathbf{x}_i$ or $\mathbf{y}_j$ is an $n$-dimensional vector, with corresponding bounder vectors $\mathbf{M}_x, \mathbf{M}_y \in \mathbb{R}^n$, we say $\mathbf{x} \preceq \mathbf{y}$ if for every dimension $k \in \{1, \ldots, n\}$, it holds that $M_x^k \leq M_y^k$, where $M_x^k$ and $M_y^k$ denote the $k$-th components of $\mathbf{M}_x$ and $\mathbf{M}_y$, respectively. Here, $\preceq$ is the hyper-rectangle order.
\end{definition}

We refer to it as the hyper-rectangle order because a minimal enclosing hyper-rectangle can be constructed based on the bounder vector to cover all token embeddings of a given sentence. Next, we discuss the relationship between the hyper-rectangle order and the convex hull. In addition, analogous to \textbf{Definition 2}, we formally define the hyper-rectangle order $\succeq$.

\begin{theorem}
Given a sequence $\mathbf{x} = (\mathbf{x}_1, \ldots, \mathbf{x}_m)$, for any subsequence of $\mathbf{x}$, denoted as $\mathbf{x}_{sub}$, it always holds that $\mathbf{x}_{sub} \preceq \mathbf{x}$.
\end{theorem}

\begin{proof}
For a given sequence $\mathbf{x} = (\mathbf{x}_1, \ldots, \mathbf{x}_m)$, suppose there is a subsequence $\mathbf{x}' = (\mathbf{x}_{i_1}, \ldots, \mathbf{x}_{i_t})$, where $\{i_1, \ldots, i_t\} \subseteq \{1, \ldots, m\}$ and $t \leq m$. Let $\mathbf{M}_x$ and $\mathbf{M}_{x'}$ be the corresponding bounder vectors of $\mathbf{x}$ and $\mathbf{x}'$, respectively. 

By the definition of bounder vector, for any dimension $k$:
\begin{align}
M_{x'}^k &= \max_{j \in \{1, \ldots, t\}} x_{i_j}^k \\
M_x^k &= \max_{i \in \{1, \ldots, m\}} x_i^k
\end{align}

Since $\{i_1, \ldots, i_t\} \subseteq \{1, \ldots, m\}$, we have:
$$\max_{j \in \{1, \ldots, t\}} x_{i_j}^k \leq \max_{i \in \{1, \ldots, m\}} x_i^k$$

Therefore, $M_{x'}^k \leq M_x^k$ for all $k$, which implies $\mathbf{x}' \preceq \mathbf{x}$.
\end{proof}

\textbf{Theorem 1} states that as the number of tokens increases, the corresponding bounder vector and its associated hyper-rectangle order also expand. This aligns with the intuition of information gain as token length grows. An exception occurs when the newly added tokens are low-content or semantically insignificant, as illustrated in the example from the Introduction. In such cases, these tokens tend to lie near the center of the distribution and, by our definition, do not affect the bounder vector.

\begin{theorem}
Given a sequence $x = (x_1, \ldots, x_m)$ and corresponding convex hull $C_x$, for any sampled sequence from $C_x$, denoted as $x_{sample}$, it always holds that $x_{sample} \preceq x$.
\end{theorem}

\begin{proof}
Since $C_x$ is the convex hull of sequence $x = (x_1, \ldots, x_m)$, for any $s \in C_x$, there exist $\lambda_i \geq 0$ with $\sum_{i=1}^m \lambda_i = 1$, such that $s = \sum_{i=1}^m \lambda_i x_i$.

Therefore, for any sequence $x_{sample} = (s_1, \ldots, s_t)$ sampled from the convex hull $C_x$, let $M_{sample}$ and $M_x$ be the corresponding bounder vectors. For any dimension $k$ and any $s_j \in x_{sample}$:
$$s_j^k = \sum_{i=1}^m \lambda_{ji} x_i^k \leq \sum_{i=1}^m \lambda_{ji} M_x^k = M_x^k$$

Thus $M_{sample}^k = \max_j s_j^k \leq M_x^k$ for all $k$, which implies $x_{sample} \preceq x$.
\end{proof}

\textbf{Theorem 2} states that, based on our definition, the convex hull of a given sequence is bounded by its corresponding bounder vector. This bounder vector can thus be used to approximate the "coverage" of the token sequence. The remaining question is: how can we effectively use it for such measurement?

\subsection{Measuring the sampling}
Here, we discuss how to use the bounded vector to measure the area covered by a given sampling result. This analysis is based on the following question: If two sampling results are drawn from the same distribution, will their corresponding bounded vectors converge as the sample size increases?

\begin{theorem}
For any two sampling results $x_a$ and $x_b$ from the same distribution (generator), with corresponding bounder vectors $M_a$ and $M_b$, for all $\epsilon > 0$, the difference between $M_a$ and $M_b$ converges to zero in probability as the sample size $m$ approaches infinity, that is: $\lim_{m \to \infty} P(\|M_a - M_b\| \geq \epsilon) = 0$.
\end{theorem}

\noindent Where $m$ is the sampling size, $P$ is the probability and $\|\cdot\|$ is the norm.

\begin{proof}
We focus on a single component $k \in \{1, \ldots, n\}$, where $n$ is the dimension of the vector space. Let $M_{a}^k$ and $M_{b}^k$ be the $k$-th components of the bounding vectors. We must show that $|M_{a}^k - M_{b}^k|$ converges to 0 in probability, i.e.,
$$\lim_{m \to \infty} P(|M_{a}^k - M_{b}^k| > \epsilon) = 0$$

Let $F_k(x)$ be the cumulative distribution function (CDF) of the $k$-th component. The CDF of the sample maximum for a sample of size $m$ is given by:
$$F_{M_{a}^k}(x) = P(M_{a}^k \leq x) = [F_k(x)]^m$$

Let $s_k = \sup\{x : F_k(x) < 1\}$ be the essential supremum of the $k$-th component.

The event $\{|M_{a}^k - M_{b}^k| > \epsilon\}$ can be split into two events: $\{M_{a}^k > M_{b}^k + \epsilon\}$ and $\{M_{b}^k > M_{a}^k + \epsilon\}$. By symmetry, we analyze the first event:

\begin{align}
&P(M_{a}^k > M_{b}^k + \epsilon) \nonumber \\
&= \int_{-\infty}^{\infty} P(M_{a}^k > y + \epsilon) f_{M_{b}^k}(y) dy \nonumber \\
&= \int_{-\infty}^{\infty} (1 - [F_k(y + \epsilon)]^m) f_{M_{b}^k}(y) dy \nonumber
\end{align}

As $m \to \infty$, the density $f_{M_{b}^k}(y)$ becomes increasingly concentrated near $s_k$. For any $y < s_k - \epsilon$, we have $F_k(y + \epsilon) < 1$, so $[F_k(y + \epsilon)]^m \to 0$. However, the mass of $f_{M_{b}^k}$ in this region vanishes. For $y$ near $s_k$, if $F_k(y + \epsilon) = 1$, then $1 - [F_k(y + \epsilon)]^m = 0$. Thus, the integral vanishes as $m \to \infty$.

By symmetry, $P(M_{b}^k > M_{a}^k + \epsilon) \to 0$ as well, so $P(|M_{a}^k - M_{b}^k| > \epsilon) \to 0$.

The probability that the norm of the difference vector exceeds $\epsilon$ is bounded by:
$$P(\|M_a - M_b\| > \epsilon) \leq \sum_{k=1}^n P(|M_{a}^k - M_{b}^k| > \epsilon/\sqrt{n})$$

Since each term goes to zero as $m \to \infty$, the entire sum also goes to zero, completing the proof.
\end{proof}

This theorem establishes a direct and rigorous link between the sample-based boundary vector and a fundamental property of the underlying distribution. The bounder vector, while a statistic derived from a finite sample, is not a random artifact. Instead, it is a consistent estimator of the distribution's true axis-aligned support.

The boundary vector captures the extremal properties of the distribution, specifically its range along each coordinate axis. As the sample size $m$ increases, the probability that the sample bounding box deviates significantly from the true boundary vector of the distribution's support becomes vanishingly small.

\subsection{Further Discussion}

In our method, we use different models for the retriever and the reader. Our primary concern is whether the summary generated by the retriever can be aligned with or accepted by the reader. Given a retrieved summary sequence, we can easily compute its corresponding bounded vector from the reader's perspective. The key question is whether this bounded vector, if treated as a proxy for the reader's output, aligns with the actual token embeddings generated by the reader. To assess this, we compare the bounded vector with the spectrum projection direction of the reader to obtain a matching score.

However, it is important to consider under what conditions our proposed framework is necessary. We assume that the token embedding distribution is sparse, and there exists a significant difference between the mean vector of the sequence $\mathbf{x} = (\mathbf{x}_1, \ldots, \mathbf{x}_n)$ and its bounded vector. In real-world scenarios, however, token embedding distributions may vary considerably. When the distribution is heavily concentrated or dense, our assumption may not hold.

To analyse this further, we adopt the commonly used assumption that token embeddings follow a multi-dimensional Gaussian distribution \cite{ng2011sparse, karvonen2025saebench}. Under this assumption, we consider the ratio between the mean vector and the bounded vector, defined as:
$$\mathcal{R} = \frac{\|\bar{\mathbf{x}}\|}{\|\mathbf{M}_x\|}$$

In most LLMs, the mean vector is close to zero. By the Strong Law of Large Numbers, the sample mean converges almost surely to the true mean vector $\boldsymbol{\mu}$:
$$\lim_{n \to \infty} \bar{\mathbf{x}} = \boldsymbol{\mu} \quad \text{(almost surely)}$$

In contrast, the bounded vector $\mathbf{M}_x$, which captures the extreme values in the sample, grows unbounded. Specifically, for a Gaussian distribution with standard deviation $\sigma$, the expected maximum of $n$ samples increases approximately as $\sigma\sqrt{2\ln n}$. Thus:
$$\lim_{n \to \infty} \|\mathbf{M}_x\| = \infty$$

Combining the asymptotic behaviours of the numerator and denominator, we find that the ratio $\mathcal{R}$ converges to zero:
$$\lim_{n \to \infty} \mathcal{R} = \lim_{n \to \infty} \frac{\|\bar{\mathbf{x}}\|}{\|\mathbf{M}_x\|} = \lim_{n \to \infty} \frac{\|\boldsymbol{\mu}\|}{\mathcal{O}(\sigma\sqrt{2\ln n})} = 0$$

This demonstrates that as the sample size increases, the mean vector becomes negligible in magnitude relative to the boundary vector.

However, in practice, the number of tokens in a summary is limited, so $n$ is bounded and $\|\mathbf{M}_x\|$ remains controlled by a function of $\sigma$. We propose using the ratio $\mathcal{R}$ as a filtering criterion to detect sparsity in token embeddings. Specifically, when $\sigma$ is large (indicating high variance and sparse distribution), the bounded vector grows large relative to the mean, resulting in a small $\mathcal{R}$. We apply our method only to summaries with sufficiently low ratios, indicating sparse and informative distributions.

\section{C. Overconfidence Analysis of Qwen and LlaMa Model}
\label{appendix:Overconfidence Analysis of Qwen}

We measured answer entropy across different models and datasets. On the HotpotQA and NQ datasets, the Qwen model tends to produce answers with very high confidence scores (i.e., predictive probabilities) compared with the LLAMA model. This observation is supported by the results in Table \ref{tab:candidate_entropy}, where we measure the entropy of generated answers across datasets. Qwen consistently yield \textit{lower entropy values}, indicating a strong tendency toward \textit{overconfident predictions}.

\begin{table}[htbp]
\centering
\begin{tabular}{lcc}
\toprule
\textbf{Model} & \textbf{HotpotQA} & \textbf{NQ}\\
\midrule
LLaMA-3.1-8B-Instruct & 1.49 & 1.16 \\
\midrule
Qwen3-8B & 0.46 & 0.38 \\
\bottomrule
\end{tabular}
\caption{Average answer entropy across datasets for different LLMs. Lower entropy indicates higher confidence in answering questions.}
\label{tab:candidate_entropy}
\end{table}

\end{document}